\documentclass[12pt]{article}
\usepackage{amsmath}
\usepackage{geometry} 
\usepackage{natbib} 
\usepackage{graphicx} 
\usepackage{subfig} 
\usepackage{fixltx2e}  
\usepackage{multicol}
\usepackage{epigraph}
\usepackage{hyperref}
\title{Does Learning Imply a Decrease in the Entropy of Behavior?}
\author{Paul Smaldino\\
Department of Anthropology\\
University of California, Davis\\
paul.smaldino@gmail.com}

\begin{document}
\maketitle
\begin{abstract}
Shannon's information entropy measures of the uncertainty of an event's outcome. If learning about a system reflects a decrease in uncertainty, then a plausible intuition is that learning should be accompanied by a decrease in the entropy of the organism's actions and/or perceptual states.  To address whether this intuition is valid, I examined an artificial organism -- a simple robot -- that learned to navigate in an arena and analyzed the entropy of the outcome variables action, state, and reward.  Entropy did indeed decrease in the initial stages of learning, but two factors complicated the scenario: (1) the introduction of new options discovered during the learning process and (2) the shifting patterns of perceptual and environmental states resulting from changes to the robot's learned movement strategies. These factors lead to a subsequent increase in entropy as the agent learned. I end with a discussion of the utility of information-based characterizations of learning.
\end{abstract}

\section{Introductory Note}
This paper is a lightly edited version of a paper I produced in 2009, as a final project for a graduate course on Natural Computation and Self Organization in department of Physics at UC Davis. Since then, I have occasionally met people who have expressed interest in the ideas presented here, which has encouraged me to return to the paper and shape it into its current form. 

The paper considers a folk definition of learning as a decrease in uncertainty, and questions whether the information entropy of an agentÕs actions or perceptual states decreases during a period of spatial learning. The method used to investigate the question involves a simulated robot, which was borrowed from another study. As such, the model system is significantly more complicated than is strictly necessary to investigate the research question at hand. That said, I believe this study is nevertheless instructive of the relationship between behavior, learning, and decision space (i.e., the set of available options). I leave it to the reader to assess its merit further.

\section{Introduction}
At the heart of adaptive behavior is learning, as past experience improves future responses to stimuli. An organism who uses fixed rules for its behavior is doomed to the evolutionary scrap yard.  For most organisms, the environment -- which includes the other organisms that make up its predators, prey, and competitors -- is too complex and uncertain to prepare for every contingency. Learning provides the solution to this uncertainty. Scenarios repeat, either exactly or approximately, and learning allows an individual to shape its responses over time toward optimal (or at least better) solutions to the problems it faces. 

In the mathematical theory of communication \citep{Shannon48}, the entity that reduces uncertainty is information, and information is measured as the reduction in entropy. Put another way, information entropy is often equated to the uncertainty inherent in a variable. An outcome known with certainty has entropy of zero. An event with many equally likely outcomes has maximal entropy. As an organism learns about its environment and moves through a process of trial and error toward a more prescribed policy of state-action pairs, it seems that we should expect the uncertainty about those encountered states and subsequent actions to decrease. 

Information theory is often applied to language learning, and it appears that in this case the entropy of actions and states {\em increases} with learning. An infant's babbling is more predictable -- with fewer possible states -- than an adult's speech. Similarly, a beginning language user relies on a few key phrases while a fluent speaker can converse on many topics. On the other hand, if one knows the statistical properties of the language in question, minimizing entropy can be quite predictive of word and sentence structure \citep{Mackay03}. The processes of language learning are quite complex, involving high level cognitive mechanisms as well as social and cultural factors. It is unclear precisely which action and state targets we should direct our analytical gaze toward, in terms of assessing uncertainty. Moreover, language is a process that is found only in humans. Instead, let us consider an arguably simpler and more general learning paradigm: spatial learning. 

Upon first encountering a new arena, a rodent moves in disorganized loops, exploring the space widely and with minimal structure to its path \citep{AvniEilam06}. Gradually, as it learns the layout of the space, it will move in direct lines between one or more landmarks. In this case, the uncertainty of the organism's actions and environmental states appears to decrease with learning. Neuroscientists have discovered much about the underlying neural architecture of spatial learning \citep{Gallistel1990, WilsonMcNaughton93, Collett98, Moser08}, and have proposed a number of models incorporating reinforcement learning and artificial neural nets to provide mechanistic explanations of the learning process \citep{BlumAbbott96, RedishTouretzky98, FosterMorrisDayan00, Erdem12}.  Such models can help advance our understanding of the relationship between brain function and well described processes. The purpose of the present exercise, however, is to assess whether learning is well described as a decrease in uncertainty. 

To address this question, I will present the results of a simple experiment using spatial learning in an artificial organism. The difficulties in studying behavior in humans and other animals are apparent. There are many, many variables that might influence learning and behavior in one way or another. For example, it is difficult to know toward what a person or animal is attending, and difficult to control for the effects of prior experience.  In contrast, consider learning in a robot.  Here the learning rules, behaviors, and internal states are exactly known. In the present study, a robot learned to navigate in a simple arena in order to maximize arbitrary movement-related rewards. The entropy of its actions and perceptual states were calculated for each phase of the learning process. The results reveal that in this simple case, learning is not well described as a decrease in entropy. Instead, the process of learning is inherently ecological, with the organism's ecology being continuously altered through the process of learning.  

\begin{figure}[h]
\begin{center}
\includegraphics[scale=0.5]{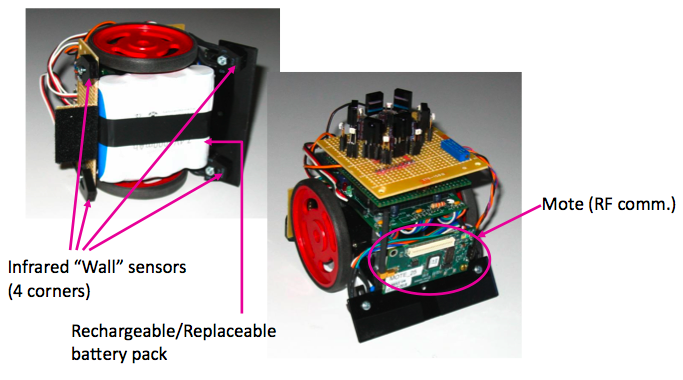}
\end{center}
\caption{The RoMADs robot, the inspiration for the simulated robot.}
\label{fig:robot}
\end{figure}

\section{Methods}
\subsection{The Model Organism and Its Environment}

The model organism was a simulated mobile robot that learned to move in a prescribed space, which was inspired by the Robotic Multi-Agent Development System (RoMADS; see Figure~\ref{fig:robot}) developed by the Dynamics of Learning Lab at UC Davis \citep{Morrison07}. The robot and its environment are intentionally very simple, in order to highlight the problems of pattern discovery that exist even in a simple system that is fully described. The present robot design was admittedly used largely out of convenience, but an advantage of such an arbitrary system is realized if it forces the readers attentions on the more general properties of learning and behavior and away from any particular naturalistic system. As this is the first study to use the present design, an arbitrarily chosen system seems appropriate. More ecologically specific systems could be fruitfully studied in the future. That said, it can also be argued that even highly unrealistic models can help us to understand important questions in biology and behavior if they are carefully considered \citep{wimsatt_1987, bedau_1999}. Such an approach is adopted here. 

Simulations rather than physical experiments were utilized out of convenience, including the need to ensure that the author had unlimited access to the study system. The robot simulation was written in Java using the MASON simulation library\footnote{The Java code is available from the author's website at \url{http://smaldino.com/wp/?attachment_id=285}, and a runnable JAR file is available at \url{http://smaldino.com/wp/?p=286}.} \citep{Luke05}.  Every attempt was made to accurately model the real robot and its environment used in the RoMADS project. Comparisons between the behaviors of the real and simulated robots showed excellent agreement. 

\begin{figure}
\begin{center}
\includegraphics[width=0.8\textwidth]{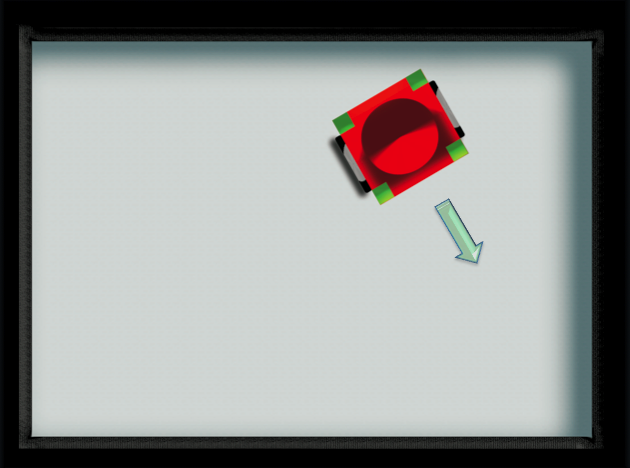}
\end{center}
\caption{A stylized view of the robot moving in its arena.}
\label{fig:robotarena}
\end{figure}

\begin{figure}
\begin{center}
\includegraphics[scale=0.3]{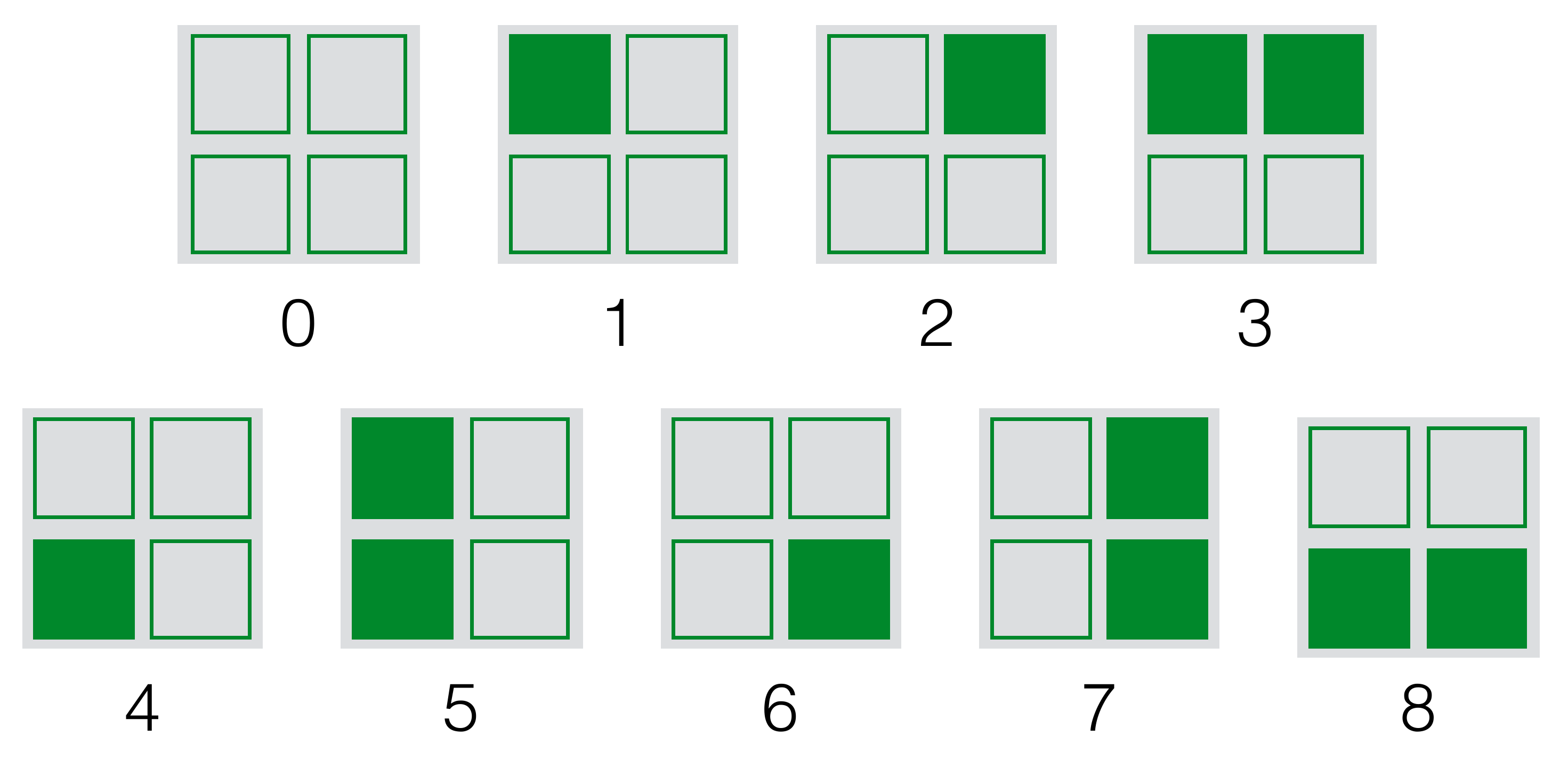}
\end{center}
\caption{The robot's possible states.}
\label{fig:robotstates}
\end{figure}

The robot began each trial in the center of a rectangular arena (Figure~\ref{fig:robotarena}), with an orientation of 0$^\circ$. The surface of the arena was white, with a black border. The white part of the arena was modeled after the physical  arena being used to train the RoMADS, and was 100 cm $\times$ 126.3 cm. The robot was a 12 cm wide square, with two wheels that allowed it to move forward and rotate, though in the simulation movement simply occurred by moving the robot forward or rotating it; the servos and wheels were not explicitly modeled. The robot had four sensors along its bottom -- one at each corner -- which detected whether the ground below them was light or dark. The robot's sensors gave rise to 16 theoretically possible {\em states}, only nine of which ever occurred in practice (Figure~\ref{fig:robotstates}). The robot's {\em actions} were drawn from a set of five possible moves: \{Forward, Left (90$^\circ$), Right (90$^\circ$), About Face (180$^\circ$), No Move\}. 

Learning was designed to maximize {\em reward}. Nonzero rewards resulted from two types of moves: forward motion, and path-clearing turns. An action of Forward resulted in an attempt to move forward for a maximum of 100 cm. If either of the front sensors detected dark, further forward movement was disallowed (i.e., if the action F was attempted, the initial state was returned, with no reward given). Forward movement was rewarded with one points per 10 cm of travel. Rewards were also received if a turning action led to a transition from a state where one or more of the front sensors were active (states 1, 2, or 3; see Figure~\ref{fig:robotstates}) to a state where both front sensors were clear. Right or left turns yielded two points, an About Face yielded one point to prevent the robot from developing a simple back and forth strategy. 

The real RoMADS robot --  the inspiration for the simulations presented here -- was constructed of metal and plastic and traveled on an imperfect surface, and thus was not always perfectly precise in its movements. These imperfections added important uncertainty to the outcomes of its actions. This was modeled in the simulation by adding noise to turns and forward motion. When moving forward, the robot's intended final position was the smaller distance of either 100 cm forward or the farthest position for which one of its front sensors was 0.1 cm past the edge of the arena. In order to model inertia and noise, this position was moved additionally forward by $x_f$ cm, where $x_f$ was a real number randomly drawn from a uniform distribution in $(0, 1.5)$. In addition, to every 90$^\circ$ or 180$^\circ$ rotation was added a noise term $x_r$, which was randomly drawn from a uniform distribution in the real interval $(-0.06, 0.06)$ radians. 

Learning was facilitated using an on-policy Monte Carlo control algorithm \citep{SuttonBarto98}. Trials were subdivided into episodes of five moves each. During the first move, the robot \emph{explored}, i.e., it chose an action at random. During each of the next four moves, the robot \emph{exploited}, i.e., it greedily picked the action known to yield the highest reward given the current state.  Policy update was possible any time the robot received a reward.  A policy is a set of state-action correspondences, in which each state leads to a specific action (which can be random). The initial policy dictated a randomly selected move in all states. 

The robot's final policy was learned over a single trial of 502 moves. This trial was deemed to be fairly typical. Importantly, the goal here was not to characterize how the robot learns about the arena {\em in general}, but rather to assess what can be determined about the current state of a learning robot by examining the intermediate phases of the learning process. More will be said about this below. Including the initial, fully random policy, a total of eight policies were used, with the eighth and final policy being settled on at the 93rd move, as shown in Table 1. The final policy was optimal in the sense that for each state-action pair, the action generated the maximum possible reward immediately following the given state. 

\begin{table}\footnotesize
\begin{center}
\begin{tabular}{ | c c c c c c c c c |}
\hline
{\bf Pol\#} & {\bf T} & {\bf P[0]}  & {\bf P[1]}  & {\bf P[2]}  & {\bf P[3]}  & {\bf P[4]}  & {\bf P[5]} & {\bf P[6]}\\ 
\hline
1	&	6	&	RAND & RAND & RAND & RAND & RAND & RAND & RAND\\
\hline
2	&	2	&	F & RAND & RAND & RAND & RAND & RAND & RAND\\
\hline
3	&	3	&	F & RAND & RAND & A & RAND & RAND & RAND\\
\hline
4	&	2	&	F & RAND & A & A & RAND & RAND & RAND\\
\hline
5	&	1	&	F & R & A & A & RAND & RAND & RAND\\
\hline
6	&	52	&	F & R & A & A & F & RAND & RAND\\
\hline
7	&	26	&	F & R & L & A & F & RAND & RAND\\
\hline
8	&	410	&	F & R & L & A & F & RAND & F\\
\hline
\end{tabular}
\caption{The policies used by the robot over the course of its learning trial. $T$ refers to the number of moves for which the robot employed the particular policy over the course of the trial run. P[$s$] is the state-action policy for being in the state $s$ (see Figure~\ref{fig:robotstates}), where the actions could be forward (F), left (L), right (R), about face (A), or random. States 7 and 8 occurred so rarely that their associated action was always RAND.}
\end{center}
\label{tab:policytable}
\end{table}

\subsection{Shannon Entropy}
Each stage of the learning process was characterized by a policy, which generated a unique pattern of behavior as the robot interacted with the environment. In order to better evaluate the behaviors generated under each policy, additional trials were simulated during which the robot employed a given policy for 200 moves without additional learning. This allowed me to acquire approximations of the frequency distributions for state, action, and reward under each policy. The inherent uncertainty in these variables was then assessed using Shannon's information entropy \citep{Shannon48}. The entropy for a variable $X$ is given by
\[
H(X) = -\sum_{x \in \mathcal{X}} p(x) \log_2 p(x)
\]
where $x$ is a realization of the variable $X$ from the alphabet $\mathcal{X}$, and $p(x)$ is the frequency of that realization. MATLAB scripts were written to calculate the entropy of the state, action, and reward variables for each policy. 

\section{Results}
\begin{figure}[h]
\begin{center}
\includegraphics[scale=.6]{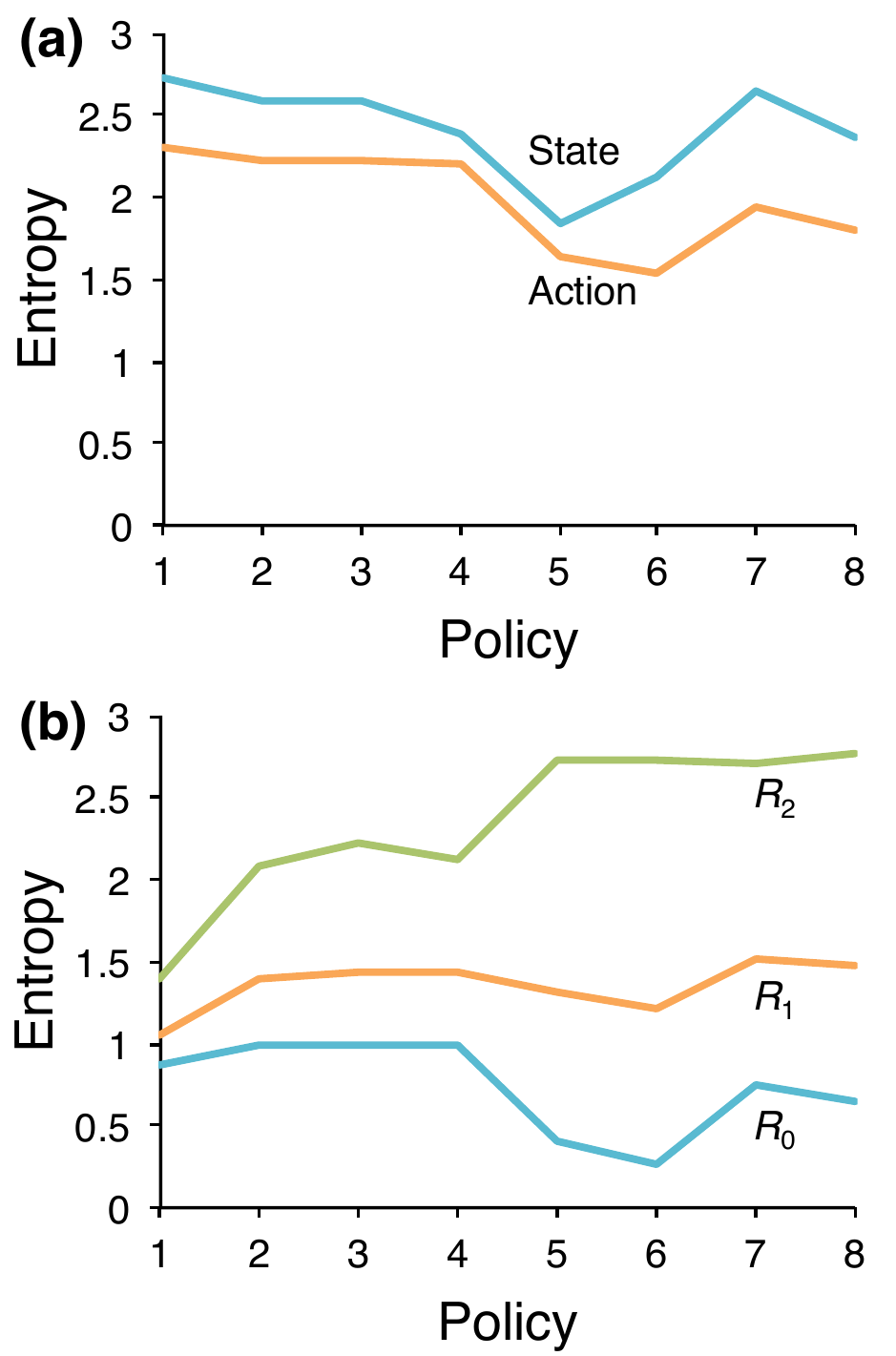}
\end{center}
\caption{Entropy for State, Action, and Reward variables.}
\label{fig:entropy1}
\end{figure}

Entropy was calculated for state, action, and reward variables under each policy (Figure~\ref{fig:entropy1}). Entropy for state and action decreased steadily between policies 1 and 5. State entropy then increased between policies 5 and 7, while action entropy increased between policies 6 and 7. The reasons why are explored below. 

\begin{figure}[h]
\begin{center}
\includegraphics[scale=0.6]{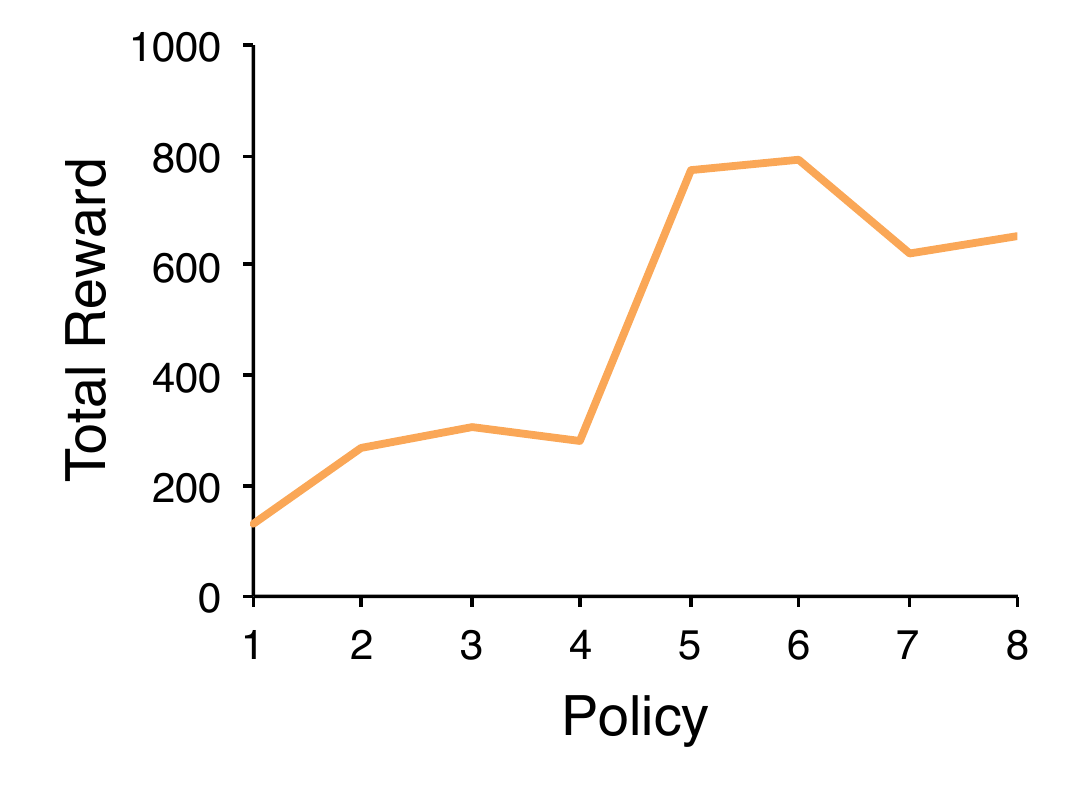}
\end{center}
\caption{The total reward in each trial of 200 time steps.}
\label{fig:rewards}
\end{figure}

Since reward ($R$) was not a discrete variable, I tested 3 different binning systems: 
\begin{align*}
R_0(\text{2 bins})&: 0, R>0\\
R_1(\text{3 bins})&: 0, 0 < R \leq 2, R>2\\
R_2(\text{11 bins})&:  0, 0 < R \leq 1, 1 < R \leq 2,2 < R \leq 3,\dots, 9<R \leq 10
\end{align*}
As seen in Figure~\ref{fig:entropy1}b, the quantitative relationship of the reward to the robot's policy varied greatly depending on how it was binned. The simplest binning, $R_0$, yields a curve very similar to that for the action variable. Under this binning, learning is characterized by the choice of actions that yield \emph{any} reward versus zero reward.  The robot's total reward during each run increases continuously as we go from policy 1 to policy 6 (Figure~\ref{fig:rewards}), which corresponds to the decrease in entropy for reward, as well as for action. When the reward is binned across 11 one-point intervals, the entropy increases dramatically starting with policy 5, the same place the reward entropy decreases when there are only 2 bins. This is due to a sudden increase in forward movement, yielding more large-valued rewards. Because these rewards values had frequency of zero in earlier trials, they did not add to the total entropy. 

It is worth noting that, contrary to what one might expect, reward was not maximal for the final adopted policy (Figure~\ref{fig:rewards}). Rather, reward declined after policy 6. Correspondingly, the entropy for action increased. It appears that learning in this case made the robot more uncertain about its actions and less rewarded. If we look at the change in policy between policies 6 and 7 (Table 1), we see that rather than going from a random action to a particular action given a state, the robot changed its action from About Face to Left when its rear right sensor was engaged. It is possible that this particular state was not encountered very often, leaving few opportunities to learn better strategies. It is also likely that Left was the better move immediately following this state (a Left turn could yield a two-point reward, while About Face only yielded one point), but that such a turn changed the larger patterns of encountered states.  Instead of allowing for a back and forth pattern with long stretches of forward motion, more 90 degree turns led to wider variety in encountered states and shorter forward periods of forward movement. This is a limitation of the learning algorithm used. Maximizing the immediate reward following a state may lead to suboptimal behavioral patterns overall. 

\begin{figure}[h]
\begin{center}
\includegraphics[scale=.6]{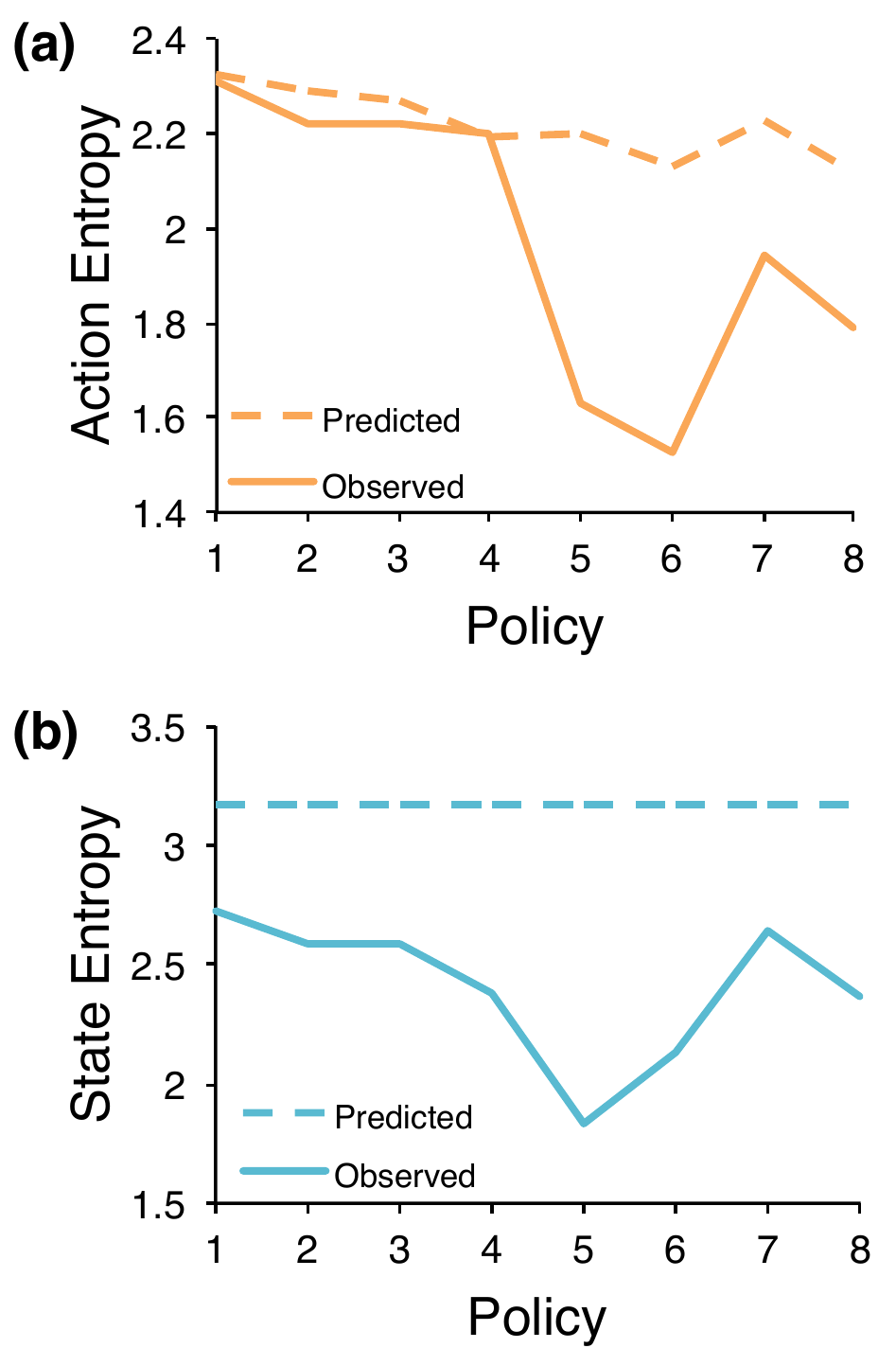}
\end{center}
\caption{Predicted entropy for policies under the assumption of uniformly distributed, randomly encountered states (dashed lines), and the observed entropy (solid lines) for action and states. The observed action entropy is very close to the predicted value in early rounds, but deviates once useful learning has occurred.}
\label{fig:entropy3}
\end{figure}

What is the relationship between the uncertainty of the robot's actions and the uncertainty of the states it encounters? Because state and action are linked, it seems possible that the entropy of the robot's actions should tell us something about the way the states are being encountered. I calculated the expected entropy for the action and state variables under each policy if the states encountered were assumed to be random and uniformly distributed. The observed action entropy was very close to the predicted values for the first four policies. A large deviation from this expectation was then observed that corresponded to the large decrease in entropy between policies 4 and 5. Without knowing anything about the robot's internal processes (its policy), we see that learning corresponds to changes in the distribution of the decision making point encountered by the robot changes. They become more structured and less random.   For the first four trials, the robot encountered states almost at random.  With the shift to policy 5, there was a large decrease in entropy. This shift from policies 4 to 5 might be thought of as more \emph{useful} learning than other shifts.  However, there was also an increase in the randomness of encountered states as the robot shifted from policy 6 to 7. This was also where the total reward dipped.  Policies 7 and 8 yielded more random state encounters than policies 5 and 6 and so they led to lower rewards, even though policy 8 optimized the immediate reward for state-action pairings.  As already noted, the way that states are encountered is a key component in calculating the utility of a given strategy, and this comes across even though the means by which state-action pairs lead to new states are not explicitly modeled by the robot's learning algorithm.

\section{Discussion}
Shannon entropy was used here to investigate the dynamics of learning in a mobile artificial organism. Based on the hypothesis that learning necessarily reflects a decrease in uncertainty, the entropies of the organism's states and actions were expected  to  progressively decrease with each policy update, reflecting improved performance through learning. Instead, performance declined in the last two policy updates, corresponding to an increase in entropy. Comparing the observed entropy with the entropy expected under an assumption of randomly encountered states illustrated a relationship between the structure of the organism's available decisions and the underlying structure of the environment. Learning does {\em not} imply a decrease in the entropy of either an individual's actions nor of its internal states. 

\begin{figure}[h]
\begin{center}
\includegraphics[scale=0.6]{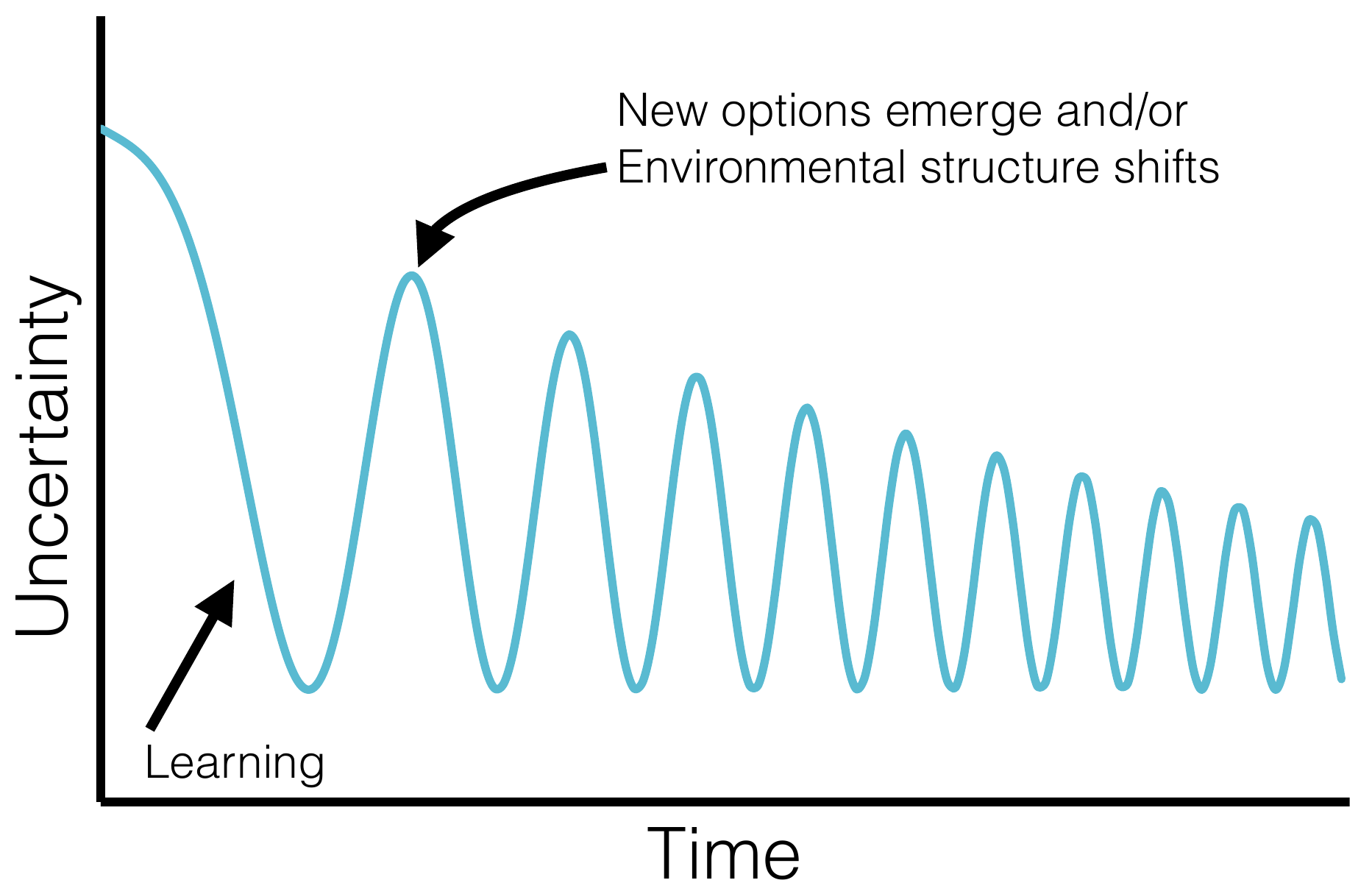}
\end{center}
\caption{A model of learning: uncertainty fluctuates as new options become available.}
\label{fig:learn}
\end{figure}

When the robot added a left turn to its repertoire in policy 7, it increased the entropy of the action and state variables and decreased its performance in terms of total reward. While this certainly reflects a limitation of the learning algorithm utilized, it may also be seen it as a metaphor for learning in general. The world is inherently uncertain, and motile organisms have evolved to make decisions in the absence of complete information. Given a new environment and little information to solve a particular problem, an organism may start by only testing a few behaviors in its repertoire, learning which ones work best in which scenarios. Gaining some level of expertise leads to new opportunities to learning, which once again increase an individual's uncertainty about the outcome for a particular scenario. More generally, new behaviors lead to new decision points.  The nature of the decision points -- the perceptual and environmental states -- generated through a behavior is directly tied to the strategies employed at previous decision points. Learn enough about your locality and you can explore further away from it, leading to new decision points and new ways to be uncertain (Figure~\ref{fig:learn}). This highlights the importance of an ecological approach to the consideration of option generation in decision making  \citep{Smaldino2012}.

There are clear limitations to the methodology employed here and to the conclusions drawn. The set of actions and states available to the robot were fully known, and large amounts of data could be obtained for each stage in the learning process.  This is an unlikely scenario for studying living animals, at least in the case of vertebrates. Thus, the simple analytic technique applied here may not be of much use in analyzing animal behavior. Nevertheless, this does not mean that more sophisticated information theoretic measures of uncertainty will not be useful in quantifying animal learning. 
In addition, the learning algorithm employed by the robot was very simple, and did not allow for multi-stage decision processes. The environment contained minimal structure and only a single individual. In the future, more sophisticated environments and learning algorithms that allow for sequences (such as temporal difference learning, \citep{SuttonBarto98}) might be employed. In that case, additional metrics such block entropy, which measures the uncertainty of sequences of moves, could be usefully employed. 

Entropy characterizes the uncertainty pertaining to a set of possible outcomes for a given event, and may be usefully applied to simply cases of organismal uncertainty \citep{Smaldino13}. However, learning in complex environments often consists of integrating and organizing information into semantic networks and action sequences \citep{Hawkins05, Hills12}. Evidence of learning might therefore be reflected not by the predictability of individual state-action pairings, but by other metrics that can better characterize the integration of complex schematic information.  Behavioral scientists often characterize learning as a decrease in the time needed to solve a repeatedly presented problem, such as navigating a maze. From a computational perspective, such an idea may correspond to Bennett's (1988) \nocite{Bennett88} concept of logical depth, in which the complexity of an algorithm is characterized by the time it takes to compute. Such an approach has some advantage over entropy approaches, because although learning really does imply less variation in behavior in some scenarios -- as in the movement of a rodent searching for a hidden platform in murky water \citep{DHooge01}, for example -- in other scenarios the opposite is true. Given a complex, open-ended problem to solve, an expert may have {\em more} variation in her behavioral output than a novice, both because she is better able to see subtle nuances in the stimuli and because she can draw on a wider range of options for behavior. 

An abstraction of organismal learning and behavior has been used to highlight a feature of learning in real organisms. In some regards, this is old news. Psychologists have long known that increased knowledge brings with it an increased awareness of the vast array of things one does not know \citep{Kruger99}. Bertrand Russell is well known for his quip that ``One of the painful things about our time is that those who feel certainty are stupid, and those with any imagination and understanding are filled with doubt and indecision" \citep{Russell51}. However, there is something more subtle at play here. Learning can change behavioral patterns. Yet learning is itself  based on previous behavioral patterns. So every time an individual learns, and that learning affects future behaviors, she potentially creates new uncertainties -- new opportunities for learning.

\begin{small}
\section*{Acknowledgments}
Many thanks to Jim Crutchfield for an inspiring course and to Benny Brown for letting me play with his robot. 
\end{small}

\bibliography{ncasoAB}{}
\bibliographystyle{apalike}
\end{document}